# Ascribing Consciousness to Artificial Intelligence

Murray Shanahan

Department of Computing
Imperial College London
180 Queen's Gate
London SW7 2RH
United Kingdom

**Abstract**

This paper critically assesses the anti-functionalist stance on consciousness adopted by certain advocates of integrated information theory (IIT), a corollary of which is that human-level artificial intelligence implemented on conventional computing hardware is necessarily not conscious. The critique draws on variations of a well-known gradual neuronal replacement thought experiment, as well as bringing out tensions in IIT's treatment of self-knowledge. The aim, though, is neither to reject IIT outright nor to champion functionalism in particular. Rather, it is suggested that both ideas have something to offer a scientific understanding of consciousness, as long as they are not dressed up as solutions to illusory metaphysical problems. As for human-level AI, we must await its development before we can decide whether or not to ascribe consciousness to it.

> "The real test is to *show* you she is a robot. Then see if you still feel she has consciousness." Alex Garland, *Ex Machina*.

## 1. Functionalism and Integrated Information

Giulio Tononi's Integrated Information Theory (IIT) has attracted widespread interest from researchers pursuing a scientific understanding of consciousness, recruiting prominent advocates, such as neuroscientist Cristoph Koch.[1] The theory, as promoted by Tononi, has both a mathematical aspect and a philosophical aspect. The mathematical aspect centres on an attempt to define a formal measure of integrated information, denoted $\Phi$, that captures the extent to which the parts of a system are influenced by the whole while retaining their distinct functionality. The philosophical aspect of the theory concerns the application of this measure to physical systems, notably to the brain, in which context it is intended to quantify the presence (or otherwise) of consciousness.

One consequence of the philosophical aspect of the theory is that functionalism is false with respect to consciousness. Indeed, IIT appears to embody a particularly strong form of anti-functionalism according to which human-level AI, if implemented on a conventional digital computer, would lack consciousness, irrespective of its high-level functional architecture and behaviour. Moreover, as made explicit in a recently published paper, Tononi and Koch believe that a neuron-by-neuron, synapse-by-synapse digital simulation of a human brain would experience nothing, even if it were behaviourally indistinguishable from the biological original.[2] Phenomenologically speaking, it would be a zombie.[3]

---

[1] Balduzzi & Tononi (2007); Oizumi, *et al*. (2014); Tononi & Koch (2015).

Not only does IIT incorporate a precisely defined metric ($\Phi$) for the "amount of consciousness", the integrated information, in a given system, it also specifies how the world is to be divided up for the purposes of applying this measure. This enables it to answer questions about the boundaries of consciousness. Does consciousness reside wholly in the brain? Is it perhaps confined to just part of the brain? Or does it encompass the rest of a person's body? Does it perhaps include parts of the environment too, such as the tools we use, as certain advocates of externalism would claim? Advocates of IIT favour the first of these options on the grounds that consciousness is intrinsic to a system of elements $x$ only if that system cannot be partitioned into a set of sub-systems any of which has a higher $\Phi$ than $x$ itself, and the brain (or perhaps some large part of it, such as the cerebral cortex) is assumed to be irreducible in this sense.

A system that is irreducible in this way is called a *complex*, and its integrated information is denoted $\Phi^{max}$. According to IIT, the brain is conscious because it includes a major complex with high $\Phi^{max}$. The theory also allows for the brain to include additional minor complexes with non-zero $\Phi^{max}$, independent sub-systems with their own minimal consciousness.[4] However, none of these sub-systems would enjoy higher $\Phi$ than the brain itself. By contrast, according to Tononi and Koch, consciousness would be absent in a digitally simulated brain because "the computer would likely not form a large complex of high $\Phi^{max}$, but break down into many mini-complexes of low $\Phi^{max}$". [5]

---

[2] Tononi & Koch (2015). Specifically, they ask "what about a computer whose software simulates in detail not just our behaviour, but even the biophysics of neurons, synapses and so on, of the relevant portion of the human brain? Could such a digital simulacrum ever be conscious?" (p.15). Their answer is that it could not. This claim is the main target of the present paper.
[3] This notion of a zombie is differs from that often employed in the philosophical literature, which is defined to be physically as well as behaviourally indistinguishable from the original.
[4] *Ibid*, fig.15(b).

To see why this might be the case, it's necessary to have some idea of how $\Phi$ is defined and deployed within the theory. Tononi & Koch offer an illustrative example of a simple system that has non-zero $\Phi$ paired with a functionally equivalent system (with more components) that has zero $\Phi$. [6] In such examples, the essential property that the system with non-zero $\Phi$ has that its zero-$\Phi$ counterpart lacks is recurrent connectivity. The presence of feedback entails that the system's transitions are irreducibly dependent on its own internal state, and that it therefore (in a specific sense) generates more information as a whole than all of its component parts generate separately. This confers a value of $\Phi$ greater than zero. The functionally equivalent feed-forward system, by contrast, generates no more information as a whole than the information generated by its (more numerous) parts, and therefore has zero $\Phi$.

On a very much larger scale, a real biological brain and its functionally equivalent digital simulation are an analogous pair of systems. The biological brain has numerous dense recurrent connections, which contribute to its high $\Phi$. By contrast, Tononi and Koch hypothesise that a digital computer will have very low $\Phi$, because it comprises (very numerous and very fast-acting) components (namely transistors) whose transitions depend on just a tiny subset of the rest of the system. This will be true whatever program the computer is running, even if that program is a flawless simulation of a human brain.

**2. Gradual Neuronal Replacement**

There is no need to elucidate this claim further in order to expose its vulnerability to an argument made popular by David Chalmers. The argument in question centres on a thought experiment in which the neurons in a person's brain are gradually replaced by electronic equivalents.[7] Because this is a thought experiment, the technological feasibility of the procedure is irrelevant. But we might imagine an injection into the bloodstream that transports tens of billions of nanobots to the brain. There, each neuron is targeted by a team of nanobots which assesses the electrochemical signalling behaviour of the cell body, maps out its axonal and dendritic connectivity, and computes the specification of a digital equivalent. The team of nanobots then reconfigures itself to match this specification. They are then ready to plug themselves in, switch out the real neuron, and take over its function.

---

[5] *Ibid*, p.15.
[6] See fig.16 of Oizumi & Tononi (2014).

A premise of the thought experiment is that replacing a neuron with a digital equivalent in this way will preserve the behaviour not only of that individual component but also of the system as a whole. This entails that the behaviour of a person whose brain undergoes a neuronal replacement will be indistinguishable from the behaviour of the fully biological original. Even that person's friends and loved ones would be unable to tell the difference. Now let us suppose that the neurons of a particular subject are replaced one at a time. The question is what happens to their consciousness as this process unfolds.

The options seem to be threefold. Either a) consciousness is an all-or-nothing property that suddenly vanishes when a certain threshold of neurons is replaced, b) consciousness gradually fades, disappearing altogether when the last neuron is replaced, or c) consciousness persists throughout the procedure. Functionalism with respect to consciousness entails the third option: persistence. By contrast, it follows directly from Tononi and Koch's argument that they favour the second option: gradual fading. According to their version of IIT, as the proportion of digitally simulated neurons gradually increases, $\Phi$ will gradually decrease, becoming negligible when the neurons are 100% replaced. In the end, the subject will become a phenomenological zombie, mindlessly insisting on its own consciousness even though there is, in fact, "no-one at home".

In order for the thought experiment to work in the context of the claim being targeted here, we need to stipulate the gradual digital replacement not only of individual neurons, but also of their connections. This is important since the thought experiment must see the graudal loss of the physical recurrent feedback connections that seem to be demanded by IIT for high Φ. So let us assume that the hypothesised nanobots, rather than taking over the function of a neuron themselves, act as tiny communication devices. The job of the nanobots then is to digitise and transmit the neuron's input signals to an external computer, where the relevant simulated computation takes place, and to receive the resulting output signal which they then pass on to other neurons. This will allow us to assume, moreover, that every time a neuron is replaced by a software-simulated equivalent, so are all the relevant structures (axons, synapses, and dendrites) connecting that neuron to others that have already undergone simulation. This will make the nanobots themselves gradually redundant as more and more of the brain is "uploaded" to the external computer. [8]

---

[7] Chalmers (1996), ch.7. See also Shanahan (2015), ch.5.

Now, let's elaborate the thought experiment a little by supposing that the person who undergoes gradual neuronal replacement is an advocate of IIT. Indeed, let's imagine that it is Tononi himself, and let's call the "person" left when all his neurons have been replaced Twin Tononi (TT). Furthermore, let's imagine that the procedure is carried out without Tononi's knowledge. (Since it is a thought experiment, we don't have to obtain ethical approval.) Recall that, by hypothesis, TT's behaviour will be indistinguishable from that of the original, fully biological Tononi. So TT will continue to advocate integrated information theory. TT will espouse the same philosophical position, rejecting functionalism on the grounds outlined in his paper with Koch. Moreover, TT will insist on his own consciousness.

The final step in the elaborated thought experiment is to expose TT to the truth. He undergoes a brain scan, and is shown what really goes on inside his own head — his biological neurons are inactive, and everything he says or does is the product of digitally simulated neuronal activity. What would TT then say about the claims made in the Tononi and Koch paper? Up to that point, of course he would have been an enthusiastic endorser of these claims, being their first author. But exposure to his true physical nature puts TT in an uncomfortable position. The options are threefold. He must either a) profess scepticism towards what he has been shown about his own inner workings, b) renounce his claim to consciousness, or c) disavow his previous rejection of functionalism.

---

[8] Chalmers (2010) outlines a similar extension to the thought experiment.

Let us set aside the first option. After all, such scepticism would be misplaced, since TT's brain is indeed a digital simulation and it would not be to his credit if no amount of evidence could persuade him of the truth. What about the second option? Would TT be willing to renounce his claim to consciousness? This seems most unlikely. Early in Tononi and Koch's paper, they assert that consciousness "is the one fact I am absolutely certain of—all the rest is conjecture". In other words, they subscribe to the near-universal philosophical view that self-knowledge about consciousness is indubitable. Now, we know that TT's inner workings are functionally equivalent to Tononi's. So whatever chains of cause and effect give rise to Tononi's assertions of indubitable self-knowledge, functionally equivalent chains of cause and effect will be at work in TT, issuing in the same pronouncements.

So we are left with the third option. Unable to deny his own consciousness, yet knowing his brain to be a digital simulation, TT will be obliged to withdraw his objection to functionalism. The question now is what the real Tononi and Koch would have to say about Twin Tononi and his retraction. If they are to retain their own anti-functionalist views, they must argue that TT is wrong, which entails that he should have accepted one of the other options. They surely wouldn't expect TT to refuse to accept the true nature of his inner workings. Rather, to maintain their anti-functionalism, they would have to assert that TT is wrong about his own consciousness. Twin Tononi, they would say, is a zombie, his most passionate protestations to the contrary notwithstanding.

## 3. The Fragility of Self-Knowledge

It would be instructive to pit Twin Tononi against real Tononi to see how they set about resolving their differences. As something of an expert on the matter, TT ought to be well placed to find flaws in Tononi and Koch's position. Indeed, we could ask the real Tononi, "How would you amend IIT if you discovered your brain was a digital simulation?" But his reply would likely be along the lines, "Well that is impossible, so the question is meaningless". Of course, if that were his reply then TT would dismiss the same question in the same way, prior to being shown the truth. We might press Tononi by asking him, "How do you know you are not a digital simulation?" And he would no doubt reply, "Because I am conscious, and according to IIT that entails that I am not a digital simulation". And again, that is exactly what TT would say too, prior to learning his true nature. Finally, we might ask Tononi, "How do you know you are conscious?" And whatever response he gave would also be the response of TT.

To get past this impasse, let's try torture.[9] First, let's imagine that the transformation from real Tononi to Twin Tononi can be reversed. Suppose that, at the flick of a switch, the digitally simulated neurons in TT's brain can be switched off, while all the dormant biological neurons are simultaneously woken up to resume their function, appropriately modified to reflect any changes (eg: in synaptic weight) undergone by their digital counterparts. According to IIT the resulting version of Tononi should be a fully conscious creature with the same high $\Phi$ as the biological original. Now suppose we carry out the following sequence of actions. We start with real Tononi and perform the procedure that turns him into TT. Then, without revealing to him his true, simulated nature, we subject TT to torture. Finally we reverse the procedure, restoring Tonini's brain to fully biological operation.

According to IIT's anti-functionalist position, TT feels nothing while he is being tortured. He screams, he bleeds, he writhes, and he begs for mercy. But all this is a sham if we accept the philosophical implications of IIT. TT has negligible $\Phi$ and therefore no consciousness. He is incapable of experiencing pain. Nevertheless, it seems a pretty safe bet that real Tononi, restored to biological function and unaware of the whole charade, will have a few complaints about his treatment. By hypothesis, his behaviour will be exactly the same as it would have been if his neurons had been biological all along. Under the circumstances, that behaviour is sure to include a grumble or two.

---

[9] In the event that Giulio Tononi reads this paper, I hope he will take all this in the spirit of intellectual enquiry that it is intended.

Now suppose we reveal to Tononi the truth. During the period of torture, his brain was merely a digital simulation, and the whole thing was just a joke. How likely is it that he would withdraw his complaints, that he would see the funny side? From Tononi's standpoint, the memory of his "fake" torture would be no less convincing than the memory of real torture. But Tononi doesn't have to compromise his intellectual position in order to complain about being tortured. "I see now that these memories aren't real", he might say. "It only seems as if I felt pain when in fact I must have felt nothing. My memories are deceiving me. Nevertheless, I am experiencing those memories, and that is horrible enough. So you really shouldn't have done that to me."

It seems implausible that anyone would go so far as to deny their own memory of torture in order to preserve their allegiance to a stance on functionalism. But fair enough. Let us allow this. By way of apology we will offer Tononi $10,000, as long as he agrees to undergo the procedure one more time. But this time we will undertake to wipe all memory of the torture from his (digital) brain prior to the switch-back to biological neurons. If the Tononi of our thought experiement really believed his anti-functionalist rhetoric, this would surely be easy money. It would involve no real pain and no unpleasant (false) memories of pain. He surely ought to agree. But I venture that no-one, not even the most dedicated (and impoverished) adherent to IIT, would agree to such a proposal.

Now let's get back to Tononi and Koch's (presumed) rejection of TT's claims to self-knowledge? Can IIT really sanction Tononi's own claim to such knowledge while denying it to his digital twin? To answer this question we need to understand what sort of mechanism might underpin a person's knowledge of their own consciousness according to IIT. The main desideratum here is that a person's consciousness must itself be causally implicated in any legitimate claim they make to this sort of self-knowledge. And according to IIT, a person is conscious because their brain has high $\Phi$. So the possession of high $\Phi$ must be causally implicated in any such claim. But how could the brain's high $\Phi$ be causally implicated in that way? How, in general, could a system have internal access to the fact of its own high $\Phi$?

Well, it is possible to imagine such a system. Suppose a system *x* is capable of computing the integrated information Φ for any given system *y*. Moreover, suppose system *x* has high Φ, and is applied to itself, yielding a correspondingly high value. This system could legitimately claim to be conscious according to IIT, and to know that it was conscious. Let's suppose this system is made of high Φ-promoting components, such as biological neurons. Now, reprising our thought experiment, imagine that we replaced all of these components with functionally equivalent digital versions, so that the resulting system was behaviourally identical to the original system *x* but had low Φ. Then, when applied to itself, this system would "do the right thing", and pronounce itself lacking in consciousness.

But the brain is not such a system. The high Φ of a person's brain plays no causal role in their pronouncements about their own consciousness. If a person claims to be certain of their own consciousness, it is not because they have studied a detailed scan of the physical structure of their brain and subjected it to the mathematical analysis demanded by IIT. A person seems to require no access to their brain's internal workings to claim indubitable knowledge of their own consciousness. It's hard to see why, from the standpoint of IIT, Tononi is any more justified in making such a claim than his digital counterpart. Tononi and TT make exactly the same pronouncements on the matter, and they do so thanks to isomorphic chains of cause and effect in which their own Φ, whether high or low, is nowhere implicated.

## 4. Science Without Metaphysics

Where does all this leave IIT? The version of the theory professed by Tononi and Koch entails a dubious position on digital brain simulation and is unable to account for the sort of self-knowledge it takes to be axiomatic. However, there is plenty to redeem it. When shorn of its metaphysical pretensions, the scientifically useful core of integrated information theory is revealed. This is the idea that a) we ascribe consciousness to humans (and other animals) because their brains are complex dynamical systems in which there is mutual influence between the whole and the parts, b) this influence can be mathematically quantified using information theory, and c) the resulting measure can be empirically validated and has clinical application.[10] There is no need to solve the "hard problem" nor to explain "qualia", notions that tend to bewitch the philosophically inclined (among whom I count myself), and lead them to see conceptual difficulties where none exist.

Let's revisit the topics of functionalism and self-knowledge with this admonition in mind. Consider functionalism. Doesn't functionalism have metaphysical pretentions of its own? Doesn't it proffer a solution to the hard problem, purport to explain qualia, and so on. Well, the aim here is not to defend functionalism *per se*. The moral, rather, is that the acceptability of any theory of consciousness depends on nothing more than how well it holds up when put to use, whether by scientists or in everyday life. So functionalism can limit its claims to the empirical: we ascribe consciousness to something when it has a certain functional organisation, and therefore exhibits certain behaviour. The required functional organisation and the right sort of behaviour might turn out to go hand in hand with the capacity to integrate information in a sense close to that of IIT. In that case a suitably modified definition of Φ would have the potential to illuminate the sort of functional organisation that underlies ascriptions of consciousness.

Be that as it may, there is no need for the functionalist or the advocate of IIT to go further, to address the "hard problem", to explain "qualia", to make metaphysical claims about what consciousness really "is". Equally, it is inappropriate to dictate, on metaphysical grounds, how a notion like integrated information is to be applied in advance of our trying it out in particular circumstances. When it comes to exotic cases, such as digital whole brain emulations, thought experiments are helpful. As for human-level artificial intelligence, whatever form it takes, whether human-like or not, we must await our encounter with it. Only then will we discover what the right attitude towards it is, whether to treat it as conscious or not.[11] We will work through our doubts, confirming them or allaying them, by observing and interacting with our creations. To stipulate in advance of this encounter what is or is not conscious is to let the metaphysical tail wag the empirical dog.

---

[10] See Seth, *et al*. (2011), for example.
[11] See Wittgenstein (1958), p.178: "'I believe that he is not an automaton' just like that, so far makes no sense. My attitude towards him is an attitude towards a soul. I am not of the *opinion* tha he has a soul".

Of course, it's all very well to attack metaphysics. But to relinquish it isn't simply a matter of linguistic censorship. For the philosophically inclined, dualistic thoughts tend to keep resurfacing, presenting a different facet each time.[12] Consider self-knowledge. It doesn't take much reflection to see that IIT's difficulties with self-knowledge also afflict functionalism. After all, the brain's lack of access to its own internal workings extends to their functional organisation. So a subject's supposedly indubitable knowledge of their own consciousness looks just as problematic for a functionalist as for a proponent of IIT. Yet no-one can doubt their own consciousness. Descartes was surely right on this point. So functionalism, like IIT, looks fatally flawed. Indeed, it's hard to see how any theory could bridge the chasm between subjective self-knowledge and objective neuroscience. So we are back to the "hard problem".[13]

However, consider the three sentences "I am conscious", "I know I am conscious", and "I am certain that I am conscious". In all three cases, the epistemological content is the same. It is nil. Epistemologically speaking, all three pronouncements are analogous to a magician stepping out of a box with hands wide and shouting "Ta da!". It would make no sense for the magician, inwardly, to ask himself "Am I sure?". Nor would it would make sense for anyone in the audience to shout out "Prove it!". The magician is presenting himself, not making a claim. Analogously, the three sentences in question stand as *expressions* of consciousness. In spite of their grammatical form, they are not propositions *about* consciousness, and they are not open to challenge as if they were.

This is not to rule out the possibility of an apparent expression of consciousness that turns out, on investigation, to be no such thing. If a mannequin with a speech synthesiser in its head announces "I am conscious" then we do not take it to be a genuine expression of consciousness. In more difficult cases, such as coma patients, extraterrestrials (if we encounter them), or human-level AIs (if we build them), the only way to decide the question is to investigate further. We might look inside their heads (if they have heads) to see what's going on there. We might observe their behaviour and note how they respond to a variety of different stimuli. Preferably we will interact with them, and the richer the interaction the more confidently will we settle on the right attitude towards them.

---

[12] See Shanahan (2010), ch.1.
[13] This, of course, is the conclusion that many philosophers have reached, including Nagel (1974) and Chalmers (1996).

This is where theories of consciousness play a role. Done properly, a theory of consciousness should address the question "Under what conditions do we ascribe consciousness to something?". This is an empirical question. To the extent that the answer takes the form "Consciousness *is* x" or "A state *x is* conscious if and only if *y*", the word "is" (the insidious copula) should be metaphysically weightless. These answers are just shorthand for "We say something is conscious when x" or "We say a state *x* is conscious when *y*". Functionalism, by these lights, labels a class of theories according to which *we describe something as* conscious when it instantiates a certain functional organisation (to be established through scientific investigation), and therefore exhibits certain behaviour.

Of course, science is embedded in human affairs, and is apt to influence the things we do and say. So the scientific study of consciousness can help us deal with tricky or exotic cases, such as coma patients or human-level artificial intelligence. So, while it might start out with the aim of describing the way we use the word "conscious" and its relatives, the scientific study of consciousness can also end up modifying or extending the way we use those words. However, nowhere in this two-way street is there any call for metaphysics.

The difficulty with IIT, as presented by Tononi and his colleagues, is that it is metaphysical through and through. It asserts that "there is an *identity* between phenomenological properties of experience and informational/causal properties of physical systems" [emphasis added].[14] To establish this identity it insists on various *a priori* postulates. For example, "of all overlapping sets of elements [of a physical system], only one set can be conscious – the one whose mechanisms specify a conceptual structure that is maximally irreducible … to independent components".[15] Based on these postulates, various controversial consequences are derived *a priori*, including (as well as a brand of panpsychism) the denial of consciousness to any human-level AI implemented on a conventional computer.

---

[14] Oizumi, *et al*. (2014), p.3.

However, it is not appropriate to pronounce on the consciousness or otherwise of human-level AI when we don't yet know what its presence in our society will be like. The temptation to do so stems from the conviction that subjective experience is a kind of "stuff", something that exists intrinsically, for itself, but whose objective character is knowable *a priori* and can be described in the language of mathematics. Of course, to deny the reality of conscious experience would be an affront to common sense. To echo Wittgenstein's words, experience *in itself* isn't a nothing, but it isn't a something either.[16] So there is no need either to affirm or deny it, and there is no call for a science of consciousness that proceeds from metaphysical assumptions or makes claims about it.

---

[15] *Ibid*.

[16] Wittgenstein (1958), §304.